\documentclass[11pt,a4paper]{article}
\usepackage[hyperref]{acl2019}
\usepackage{times}
\usepackage{latexsym}
\usepackage{multirow}
\usepackage{graphicx}
\usepackage{amsmath}
\usepackage{amssymb}
\usepackage{amsthm}
\usepackage{booktabs}
\usepackage{adjustbox}
\usepackage{url}

\makeatletter
\DeclareRobustCommand{\cev}[1]{%
  {\mathpalette\do@cev{#1}}%
}
\newcommand{\do@cev}[2]{%
  \vbox{\offinterlineskip
    \sbox\z@{$\m@th#1 x$}%
    \ialign{##\cr
      \hidewidth\reflectbox{$\m@th#1\vec{}\mkern4mu$}\hidewidth\cr
      \noalign{\kern-\ht\z@}
      $\m@th#1#2$\cr
    }%
  }%
}
\makeatother

\aclfinalcopy 
\def\aclpaperid{83} 

\title{Learning Question-Guided Video Representation for \\
Multi-Turn Video Question Answering}

\author{Guan-Lin Chao\\
  Carnegie Mellon University\\
  \texttt{guanlinchao@cmu.edu} \\\And
  Abhinav Rastogi \\
  Google AI \\
  \texttt{abhirast@google.com} \\\And
  Semih Yavuz \\
  University of California, Santa Barbara \\
  \texttt{syavuz@cs.ucsb.edu} \\\AND
  Dilek Hakkani-T\"{u}r \\
  Amazon Alexa AI \\
  \texttt{dilek@ieee.org} \\\And
  Jindong Chen \\
  Google AI \\
  \texttt{jdchen@google.com} \\\And
  Ian Lane \\
  Carnegie Mellon University \\
  \texttt{lane@cmu.edu}
  }

\date{}

\begin{document}
\maketitle

\begin{abstract}
Understanding and conversing about dynamic scenes is one of the key capabilities of AI agents that navigate the environment and convey useful information to humans. Video question answering is a specific scenario of such AI-human interaction where an agent generates a natural language response to a question regarding the video of a dynamic scene. Incorporating features from multiple modalities, which often provide supplementary information, is one of the challenging aspects of video question answering. Furthermore, a question often concerns only a small segment of the video, hence encoding the entire video sequence using a recurrent neural network is not computationally efficient. Our proposed question-guided video representation module efficiently generates the token-level video summary guided by each word in the question. The learned representations are then fused with the question to generate the answer.
Through empirical evaluation on the Audio Visual Scene-aware Dialog (AVSD) dataset~\cite{alamri2019audiovisual}, our proposed models in single-turn and multi-turn question answering achieve state-of-the-art performance on several automatic natural language generation evaluation metrics.
\end{abstract}
\section{Introduction}
\begin{figure}[t]
    \includegraphics[width=\linewidth]{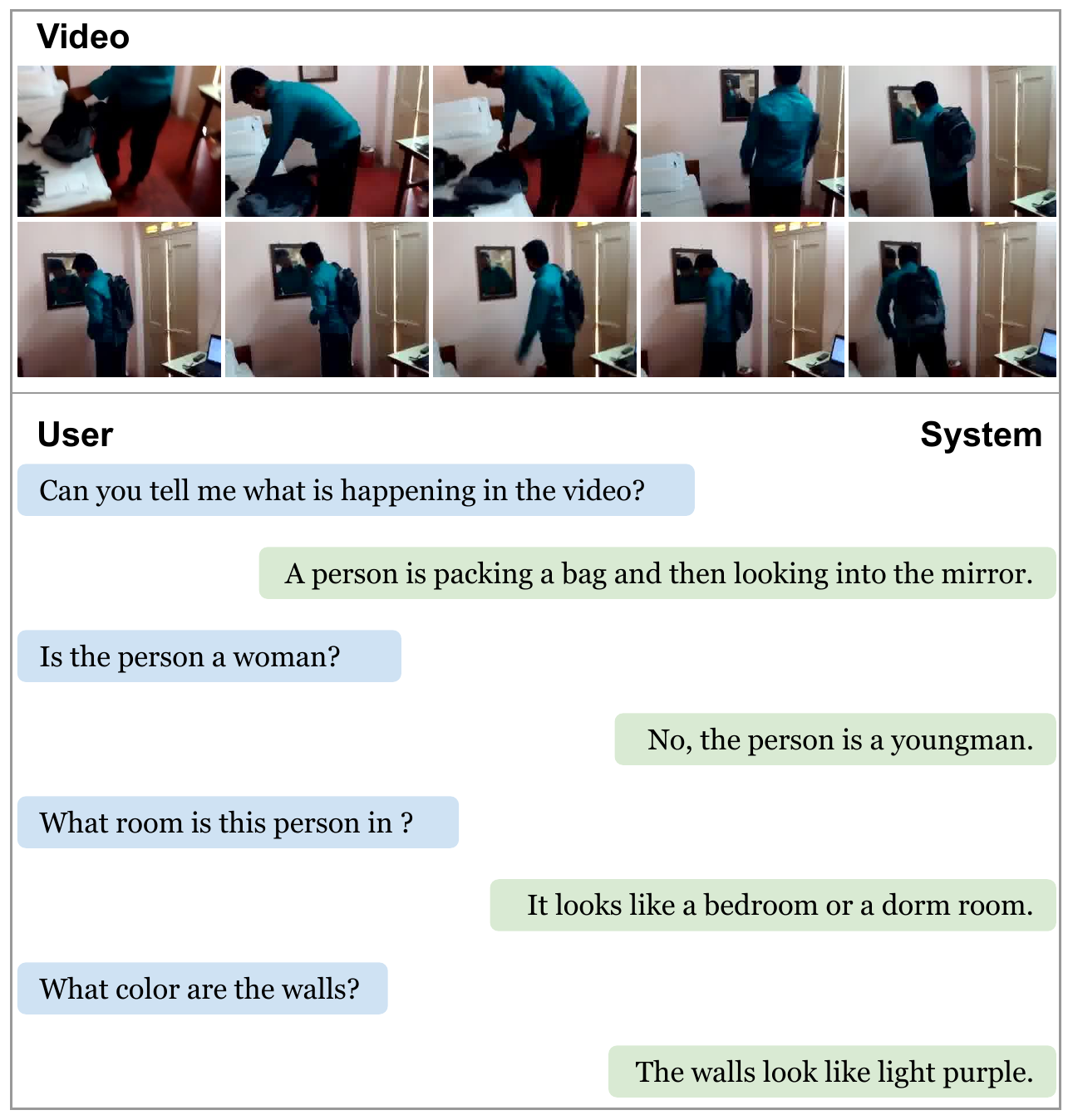}
    \caption{An example from the AVSD dataset.
    Each example contains a video and its associated question answering dialogue regarding the video scene.
    }
    \label{fig:problem}
\end{figure}

Nowadays dialogue systems are becoming more and more ubiquitous in our lives. It is essential for such systems to perceive the environment, gather data and convey useful information to humans in an accessible fashion. Video question answering (VideoQA) systems provide a convenient way for humans to acquire visual information about the environment. If a user wants to obtain information about a dynamic scene, one can simply ask the VideoQA system a question in natural language, and the system generates a natural-language answer.
The task of a VideoQA dialogue system in this paper is described as follows. Given a video as grounding evidence, in each dialogue turn, the system is presented a question and is required to generate an answer in natural language.
Figure~\ref{fig:problem} shows an example of multi-turn VideoQA. It is composed of a video clip and a dialogue, where the dialogue contains open-ended question answer pairs regarding the scene in the video. In order to answer the questions correctly, the system needs to be effective at understanding the question, the video and the dialogue context altogether.

Recent work on VideoQA has shown promising performance using multi-modal attention fusion for combination of features from different modalities~\cite{xu2017video, zeng2017leveraging, zhao2018open, gao2018motion}. However, one of the challenges is that the length of the video sequence can be very long and the question may concern only a small segment in the video.
Therefore, it may be time inefficient to encode the entire video sequence using a recurrent neural network.

In this work, we present the question-guided video representation module which learns 1) to summarize the video frame features efficiently using an attention mechanism and 2) to perform feature selection through a gating mechanism.
The learned question-guided video representation is a compact video summary for each token in the question.
The video summary and question information are then fused to create multi-modal representations. The multi-modal representations and the dialogue context are then passed as input to a sequence-to-sequence model with attention to generate the answer (Section~\ref{sec:approach}).
We empirically demonstrate the effectiveness of the proposed methods using the AVSD dataset~\cite{alamri2019audiovisual} for evaluation (Section~\ref{sec:experiments}). The experiments show that our model for single-turn VideoQA achieves state-of-the-art performance, and our multi-turn VideoQA model shows competitive performance, in comparison with existing approaches (Section~\ref{sec:results}).
\begin{figure*}[!t]
    \center
	\includegraphics[width=\textwidth]{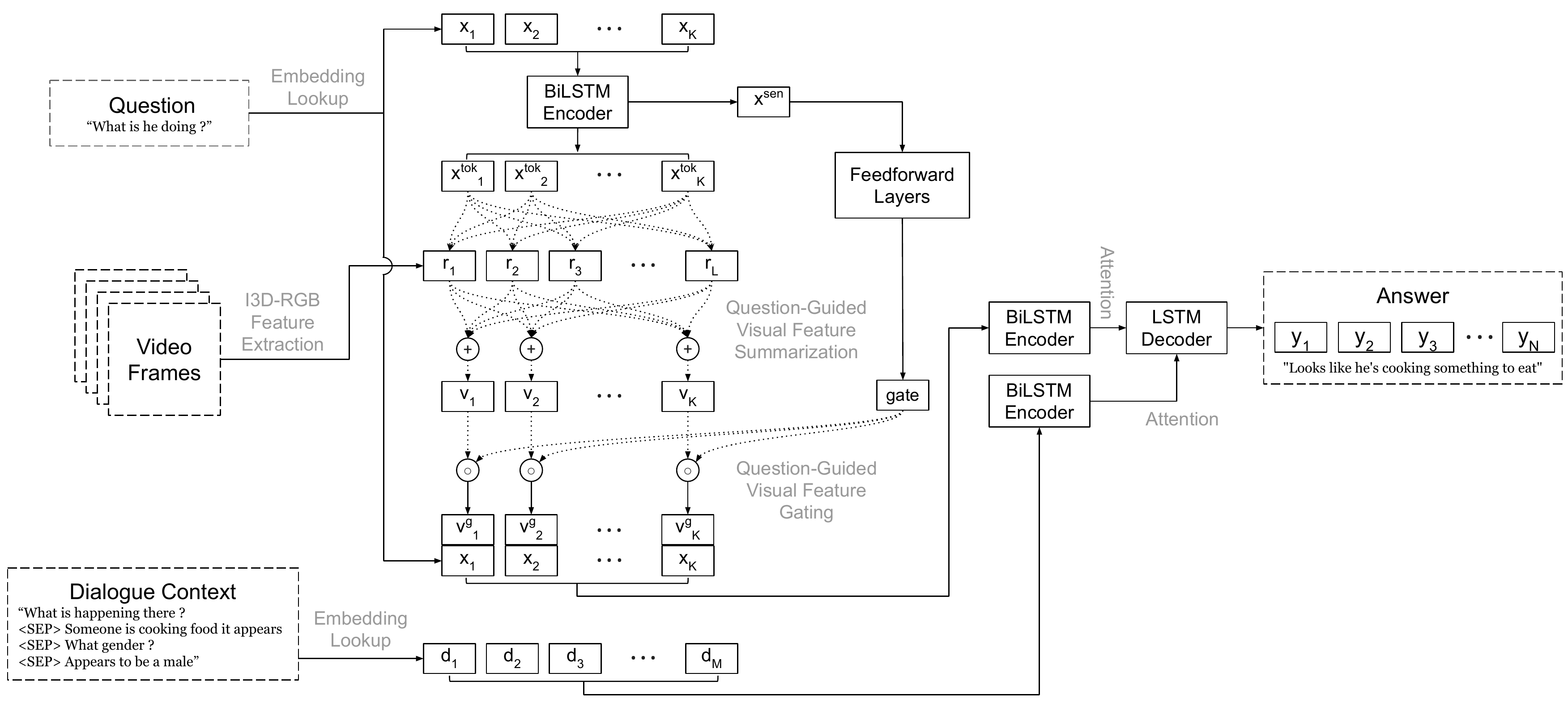}
	\caption{Overview of the proposed approach. First the I3D-RGB frame features are extracted. The question-guided video representation module takes as input the question sentence and the I3D-RGB features, generates a video representation for each token and applies gating using question as guidance. Then the question tokens are augmented by the per-token video representations and encoded by a bidirectional LSTM encoder. Similarly, the dialogue context is encoded by a bidirectional LSTM encoder. Finally, the LSTM answer decoder predicts the answer sequence.}
	\label{fig:model-overview}
\end{figure*}

\section{Related Work}
In the recent years, research on visual question answering has accelerated following the release of multiple publicly available datasets. These datasets include COCO-QA~\cite{ren2015exploring}, VQA~\cite{agrawal2017vqa}, and Visual Madlibs~\cite{yu2015visual} for image question answering and MovieQA~\cite{tapaswi2016movieqa}, TGIF-QA~\cite{jang2017tgif}, and TVQA~\cite{lei2018tvqa} for video question answering.

\subsection{Image Question Answering}
The goal of image question answering is to infer the correct answer, given a natural language question related to the visual content of an image. It assesses the system's capability of multi-modal understanding and reasoning regarding multiple aspects of humans and objects, such as their appearance, counting, relationships and interactions~\cite{lei2018tvqa}.
State-of-the-art image question answering models make use of spatial attention to obtain a fixed length question-dependent embedded representation of the image, which is then combined with the question feature to predict the answer~\cite{yang2016stacked, xu2016ask, kazemi2017show, anderson2018bottom}.
Dynamic memory~\cite{kumar2016ask, xiong2016dynamic} and co-attention mechanism~\cite{lu2016hierarchical, ma2018visual} are also adopted to model sophisticated cross-modality interactions.

\subsection{Video Question Answering}
VideoQA is a more complex task. As a video is a sequence of images, it contains not only appearance information but also motion and transitions. Therefore, VideoQA requires spatial and temporal aggregation of image features to encode the video into a question-relevant representation. 
Hence, temporal frame-level attention is utilized to model the temporal dynamics, where frame-level attribute detection and unified video representation are learned jointly~\cite{ye2017video, xu2017video, mun2017marioqa}.
Similarly, ~\citeauthor{lei2018tvqa}~\shortcite{lei2018tvqa} use Faster R-CNN ~\cite{ren2015faster} trained with the Visual Genome~\cite{krishna2017visual} dataset to detect object and attribute regions in each frame, which are used as input features to the question answering model.
Previous works also adopt various forms of external memory~\cite{sukhbaatar2015end, kumar2016ask, graves2016hybrid} to store question information, which allows multiple iterations of question-conditioned inference on the video features~\cite{na2017read, kim2017deepstory, zeng2017leveraging, gao2018motion, fan2019heterogeneous}.

\subsection{Video Question Answering Dialogue}
Recently in DSTC7,~\citeauthor{alamri2019audiovisual} \shortcite{alamri2019audiovisual} introduce the Audio-Visual Scene-aware Dialog (AVSD) dataset for multi-turn VideoQA. In addition to the challenge of integrating the questions and the dynamic scene information, the dialogue system also needs to effectively incorporate the dialogue context for coreference resolution to fully understand the user's questions across turns. To this end,~\citeauthor{alamri@DSTC7}~\shortcite{alamri@DSTC7} use two-stream inflated 3D ConvNet (I3D) model~\cite{carreira2017quo} to extract spatiotemporal visual frame features (I3D-RGB features for RGB input and I3D-flow features for optical flow input), and propose the Na\"ive Fusion method to combine multi-modal inputs based on the hierarchical recurrent encoder (HRE) architecture~\cite{das2017visual}.
~\citeauthor{hori2018end}~\shortcite{hori2018end} extend the Na\"ive Fusion approach and propose the Attentional Fusion method which learns multi-modal attention weights to fuse features from different modalities. 
~\citeauthor{zhuang2019investigation}~\shortcite{zhuang2019investigation} modify the Attentional Fusion method and propose to use Maximum Mutual Information (MMI)~\cite{bahl1986maximum} as the training objective.
Besides the HRE architecture, the multi-source sequence-to-sequence (Multi-Source Seq2Seq) architecture with attention~\cite{zoph2016multi, firat2016multi} is also commonly applied~\cite{pasunuru2019dstc7, kumar2019context, yeh2019reactive}.
Previous works~\cite{sanabria2019cmu, le2019end, pasunuru2019dstc7} also explore various attention mechanisms to incorporate the different modal inputs, such as hierarchical attention~\cite{libovicky2017attention} and cross attention~\cite{seo2017bidirectional}.
For modeling visual features, ~\citeauthor{lin2019entropy}~\shortcite{lin2019entropy} propose to use Dynamic memory networks~\cite{kumar2016ask} and ~\citeauthor{nguyen2019from}~\shortcite{nguyen2019from} propose to use feature-wise linear modulation layers~\cite{perez2018film}.
\section{Approach}
\label{sec:approach}
We formulate the multi-turn VideoQA task as follows.
Given a sequence of raw video frames $\mathbf{f}$, the embedded question sentence $\mathbf{x} = \{x_1, \ldots, x_K\}$ and the single concatenated embedded sentence of the dialogue context $\mathbf{d} = \{d_1, \ldots, d_M\}$, the output is an answer sentence $\mathbf{y} = \{y_1, \ldots, y_N\}$.

The architecture of our proposed approach is illustrated in Figure~\ref{fig:model-overview}.
First the Video Frame Feature Extraction Module extracts the I3D-RGB frame features from the video frames (Section~\ref{sec:i3d_rgb_extraction}).
The Question-Guided Video Representation Module takes as input the embedded question sentence and the I3D-RGB features, and generates a compact video representation for each token in the question sentence (Section~\ref{sec:question_guided_video_representation}). In the Video-Augmented Question Encoder, the question tokens are first augmented by their corresponding per-token video representations and then encoded by a bidirectional LSTM (Section~\ref{sec:video_augmented_question_encoder}). 
Similarly, in the Dialogue Context Encoder, the dialogue context is encoded by a bidirectional LSTM (Section~\ref{sec:dialogue_context_encoder}).
Finally, in the Answer Decoder, the outputs from the Video-Augmented Question Encoder and the Dialogue Context Encoder are used as attention memory for the LSTM decoder to predict the answer sentence (Section~\ref{sec:answer_decoder}). Our encoders and decoder work in the same way as the multi-source sequence-to-sequence models with attention~\cite{zoph2016multi, firat2016multi}.

\subsection{Video Frame Feature Extraction Module}
\label{sec:i3d_rgb_extraction}
In this work, we make use of the I3D-RGB frame features as the visual modality input, which are pre-extracted and provided in the AVSD dataset~\cite{alamri2019audiovisual}. Here we briefly describe the I3D-RGB feature extraction process, and we refer the readers to~\cite{carreira2017quo} for more details of the I3D model. Two-stream Inflated 3D ConvNet (I3D) is a state-of-the-art action recognition model which operates on video inputs. The I3D model takes as input two streams of video frames: RGB frames and optical flow frames. The two streams are separately passed to a respective 3D ConvNet, which is inflated from 2D ConvNets to incorporate the temporal dimension. Two sequences of spatiotemporal features are produced by the respective 3D ConvNet, which are jointly used to predict the action class.
The I3D-RGB features provided in the AVSD dataset are intermediate spatiotemporal representations from the "Mixed\_5c" layer of the RGB stream's 3D ConvNet.
The AVSD dataset uses the I3D model parameters pre-trained on the Kinetics dataset~\cite{kay2017kinetics}.
To reduce the number of parameters in our model, we use a trainable linear projection layer to reduce the dimensionality of I3D-RGB features from 2048 to 256. Extracted from the video frames $\mathbf{f}$ and projected to a lower dimension, the sequence of dimension-reduced I3D-RGB frame features are denoted by $\mathbf{r} = \{r_1, \ldots, r_L\}$, where $r_i \in \mathrm{R}^{256}, \forall i$.

\subsection{Question-Guided Video Representation Module}
\label{sec:question_guided_video_representation}
We use a bidirectional LSTM network to encode the sequence of question token embedding $\mathbf{x} = \{x_1, \ldots, x_K\}$. The token-level intermediate representations are denoted by $\mathbf{x}^\text{tok} = \{x^\text{tok}_1, \ldots, x^\text{tok}_K\}$, and the embedded representation of the entire question is denoted by $x^\text{sen}$. These outputs will be used to guide the video representation.
\begin{align}
    &\vec{h}_0 = \cev{h}_{K+1} = \mathbf{0} \\
    &\hspace{5mm}\vec{h}_k = \textrm{LSTM}^\text{forw}_\text{guide}(x_k, \vec{h}_{k - 1}) \\
    &\hspace{5mm}\cev{h}_k = \textrm{LSTM}^\text{back}_\text{guide}(x_k, \cev{h}_{k + 1}) \\
    &\hspace{5mm}x^\text{tok}_k = \vec{h}_k \oplus \cev{h}_k \\
    &\hspace{5mm} \forall k \in \{1, \ldots, K\} \notag \\
    &x^\text{sen} = \vec{h}_K \oplus \cev{h}_1
\end{align}
where $\oplus$ denotes vector concatenation; $\vec{\mathbf{h}}$ and $\cev{\mathbf{h}}$ represent the local forward and backward LSTM hidden states.

\subsubsection{Per-Token Visual Feature Summarization}
Generally the sequence length of the video frame features is quite large, as shown in Table \ref{tab:dstc7_stat}. Therefore it is not computationally efficient to encode the video features using a recurrent neural network. We propose to use the attention mechanism to generate a context vector to efficiently summarize the I3D-RGB features.
We use the trilinear function~\cite{seo2017bidirectional} as a similarity measure to identify the frames most similar to the question tokens. For each question token $x_k$, we compute the similarity scores of its encoded representation $x^\text{tok}_k$ with each of the I3D-RGB features $\mathbf{r}$. The similarity scores $\mathbf{s}_k$ are converted to an attention distribution $\mathbf{w}^\text{att}_k$ over the I3D-RGB features by the softmax function. And the video summary $v_k$ corresponding to the question token $x_k$ is defined as the attention weighted linear combination of the I3D-RGB features. We also explored using dot product for computing similarity and empirically found out it yields suboptimal results.
\begin{align}
    s_{k, l} &= \textrm{trilinear} (x^\text{tok}_k, r_l) \\
    &= W_\text{sim} [x^\text{tok}_k \oplus r_l \oplus (x^\text{tok}_k \odot r_l)] \\
    &\hspace{5mm} \forall l \in \{1, \ldots, L\} \notag \\
    \textbf{w}^\text{att}_k &= \textrm{softmax}(\mathbf{s}_k) \\
    v_k &= \sum_{l = 1}^L w^\text{att}_{k,l}~r_l \\
    \forall k &\in \{1, \ldots, K\} \notag
\end{align}
where $\odot$ denotes element-wise multiplication, and $W_\text{sim}$ is a trainable variable.

\subsubsection{Visual Feature Gating}
Not all details in the video are important for answering a question. Attention helps in discarding the unimportant frames in the time dimension. We propose a gating mechanism which enables us to perform feature selection within each frame. We project the sentence-level question representation $x^\text{sen}$ through fully-connected layers with ReLU nonlinearity to generate a gate vector $g$. For each question token $x_k$, its corresponding video summary $v_k$ is then multiplied element-wise with the gate vector $g$ to generate a gated visual summary $v^\text{g}_k$. We also experimented applying gating on the dimension-reduced I3D-RGB features $\mathbf{r}$, prior to the per-token visual feature summarization step, but it resulted in an inferior performance.
\begin{align}
    g &= \textrm{sigmoid}(W_\text{g, 1}(\textrm{ReLU}(W_\text{g, 2} x^\text{sen} + b_\text{g, 2}) \notag\\
    &\hspace{20mm}+ b_\text{g, 1}) \\
    v^g_k &= v_k \odot g \\
    \forall k &\in \{1, \ldots, K\} \notag
\end{align}
where $W_\text{g, 1}$, $b_\text{g, 1}$, $W_\text{g, 2}$, $b_\text{g, 2}$ are trainable variables.

\subsection{Video-Augmented Question Encoder}
\label{sec:video_augmented_question_encoder}
Given the sequence of per-token gated visual summary $\mathbf{v}^\text{g} = \{v^\text{g}_1, \ldots, v^\text{g}_K\}$, we augment the question features by concatenating the embedded question tokens $\mathbf{x} = \{x_1, \ldots, x_K\}$ with their associated per-token video summary. The augmented question features are then encoded using a bidirectional LSTM. The token-level video-augmented question features are denoted by $\mathbf{q}^\text{tok} = \{q^\text{tok}_1, \ldots, q^\text{tok}_K\}$, and the sentence-level feature is denoted by $q^\text{sen}$.
\begin{align}
    &\vec{h}_0 = \cev{h}_{K+1} = \mathbf{0} \\
    &\hspace{5mm}\vec{h}_k = \textrm{LSTM}^\text{forw}_\text{ques}(x_k \oplus v^\text{g}_k, \vec{h}_{k - 1}) \\
    &\hspace{5mm}\cev{h}_k = \textrm{LSTM}^\text{back}_\text{ques}(x_k \oplus v^\text{g}_k, \cev{h}_{k + 1}) \\
    &\hspace{5mm}q^\text{tok}_k = \vec{h}_k \oplus \cev{h}_k \\
    &\hspace{5mm} \forall k \in \{1, \ldots, K\} \notag \\
    &q^\text{sen} = \vec{h}_K \oplus \cev{h}_1
\end{align}
where $\vec{\mathbf{h}}$ and $\cev{\mathbf{h}}$ represent the local forward and backward LSTM hidden states.

\subsection{Dialogue Context Encoder}
\label{sec:dialogue_context_encoder}
Similar to the video-augmented question encoder, we encode the embedded dialogue context tokens $\mathbf{d} = \{d_1, \ldots, d_M\}$ using a bidirectional LSTM. The embedded token-level representations are denoted by $\mathbf{d}^\text{tok} = \{d^\text{tok}_1, \ldots, d^\text{tok}_M\}$.
\begin{align}
    &\vec{h}_0 = \cev{h}_{M+1} = \mathbf{0} \\
    &\hspace{5mm}\vec{h}_m = \textrm{LSTM}^\text{forw}_\text{dial}(d_m, \vec{h}_{m - 1}) \\
    &\hspace{5mm}\cev{h}_m = \textrm{LSTM}^\text{back}_\text{dial}(d_m, \cev{h}_{m + 1}) \\
    &\hspace{5mm}d^\text{tok}_m = \vec{h}_m \oplus \cev{h}_m \\
    &\hspace{5mm} \forall m \in \{1, \ldots, M\} \notag
\end{align}
where $\vec{\mathbf{h}}$ and $\cev{\mathbf{h}}$ represent the local forward and backward LSTM hidden states.

\subsection{Answer Decoder}
\label{sec:answer_decoder}
The final states of the forward and backward LSTM units of the question encoder are used to initialize the state of answer decoder. 
Let $y_n$ be the output of the decoder at step $n$, where $1 \leq n \leq N$, $y_0$ be the special start of sentence token and $y^\text{emb}_{n}$ be the embedded representation of $y_n$.
At a decoder step $n$, the previous decoder hidden state $h_{n-1}$ is used to attend over $\mathbf{q}^\text{tok}$ and $\mathbf{d}^\text{tok}$ to get the attention vectors $h^\text{att, q}_n$ and $h^\text{att, d}_n$ respectively.
These two vectors retrieve the relevant features from the intermediate representations of the video-augmented question encoder and the dialogue context encoder, both of which are useful for generating the next token of the answer. At each decoder step, the decoder hidden state $h_n$ is used to generate a distribution over the vocabulary. The decoder output $y_n^*$ is defined to be $\textrm{argmax}_{y_n}~p(y_n | y_{\leq n-1})$.

\begin{align}
    &h_0 = q^\text{sen} \\
    &\hspace{4mm}s^\text{q}_{n, k} = v_\text{ans, q}^\top~\textrm{tanh}(W_\text{ans, q} [h_{n-1} \oplus q^\text{tok}_k]) \\
    &\hspace{4mm} \forall k \in \{1, \ldots, K\} \notag \\
    &\hspace{2mm}\textbf{w}^\text{q}_n = \textrm{softmax}(\mathbf{s}^\text{q}_n) \\
    &\hspace{2mm}h^\text{att, q}_n = \sum_{k = 1}^K w^\text{q}_{n, k} ~q^\text{tok}_k \\
    &\hspace{4mm}s^\text{d}_{n, m} = v_\text{ans, d}^\top~\textrm{tanh}(W_\text{ans, d} [h_{n-1} \oplus d^\text{tok}_m]) \\
    &\hspace{4mm} \forall m \in \{1, \ldots, M\} \notag \\
    &\hspace{2mm}\textbf{w}^\text{d}_n = \textrm{softmax}(\mathbf{s}^\text{d}_n) \\
    &\hspace{2mm}h^\text{att, d}_n = \sum_{m = 1}^M w^\text{d}_{n, m} ~d^\text{tok}_m \\
    &\hspace{2mm}h_n = \textrm{LSTM}_\text{ans}(y^\text{emb}_{n-1}, [h^\text{att, q}_n \oplus h^\text{att, d}_n \oplus h_{n-1}]) \\
    &\hspace{2mm}p(y_n | y_{\leq n-1}) = \textrm{softmax}(W_\text{ans} h_n + b_\text{ans}) \\
    &\hspace{2mm} \forall n \in \{1, \ldots, N\} \notag
\end{align}
where $\mathbf{h}$ represents the local LSTM hidden states, and $W_\text{ans, q}$, $W_\text{ans, d}$, $W_\text{ans}$, $b_\text{ans}$ are trainable variables.
\section{Experiments}
\label{sec:experiments}

\begin{table}
\begin{adjustbox}{max width=\linewidth}
  \begin{tabular}{ l | r  r  r }
    \toprule
     & Training & Validation & Test \\
    \midrule
    \# of dialogues & 7659 & 732 & 733 \\
    \hline
    \# of turns & 153,180 & 14,680 & 14,660 \\
    \hline
    \# of words & 1,450,754 & 138,314 & 138,790 \\
    \hline
    Avg. length of & \multirow{2}{*}{8.5} & \multirow{2}{*}{8.4} & \multirow{2}{*}{8.5} \\
    question ($K$) \\
    \hline
    Avg. length of & \multirow{2}{*}{179.2} & \multirow{2}{*}{173.0} & \multirow{2}{*}{171.3} \\
    I3D-RGB ($L$)\\
    \bottomrule
  \end{tabular}
\end{adjustbox}
\caption{Data statistics of the AVSD dataset. We use the official training set, and the public (\textit{i.e.,} prototype) validation and test sets. We also present the average length of the question token sequences and the I3D-RGB frame feature sequences to highlight the importance of time efficient video encoding without using a recurrent neural network. The sequence lengths of the questions and I3D-RGB frame features are denoted by $K$ and $L$ respectively in the model description (Section~\ref{sec:approach}).}
\label{tab:dstc7_stat}
\end{table}

\subsection{Dataset}
We consider the Audio-Visual Scene-aware Dialog (AVSD) dataset~\cite{alamri2019audiovisual} for evaluating our proposed model in single-turn and multi-turn VideoQA. We use the official release of train set for training, and the public (\textit{i.e.,} prototype) validation and test sets for inference.
The AVSD dataset is a collection of text-based human-human question answering dialogues based on the video clips from the CHARADES dataset~\cite{sigurdsson2016hollywood}.
The CHARADES dataset contains video clips of daily indoor human activities, originally purposed for research in video activity classification and localization. Along with the video clips and associated question answering dialogues, the AVSD dataset also provides the pre-extracted I3D-RGB visual frame features using a pre-trained two-stream inflated 3D ConvNet (I3D) model~\cite{carreira2017quo}. The pre-trained I3D model was trained on the Kinetics dataset~\cite{kay2017kinetics} for human action recognition.

\begin{table*}[t!]
\centering
\begin{adjustbox}{max width=\textwidth}
 \begin{tabular}{l | l l l l l l l }
 \toprule
 \textbf{Single-Turn VideoQA Models} & BLEU-1 & BLEU-2 & BLEU-3 & BLEU-4 & METEOR & ROUGE-L & CIDEr \\
 \midrule
 Na\"ive Fusion & 27.7 & 17.5 & 11.8 & 8.3 & 11.7 & 28.8 & 74.0 \\
 Multi-source Seq2Seq & - & - & - & 8.83 & 12.43 & 34.23 & 95.54 \\
 Ours & \textbf{29.56}$\pm$0.75 & \textbf{18.60}$\pm$0.49 & \textbf{13.16}$\pm$0.33 & \textbf{9.77}$\pm$0.21 & \textbf{13.19}$\pm$0.20 & \textbf{34.29}$\pm$0.19 & \textbf{101.75}$\pm$1.03\\
 \bottomrule
 \toprule
 \textbf{Multi-Turn VideoQA Models} & BLEU-1 & BLEU-2 & BLEU-3 & BLEU-4 & METEOR & ROUGE-L & CIDEr \\
 \midrule
 Na\"ive Fusion & 27.7 & 17.6 & 12.0 & 8.5 & 11.8 & 29.0 & 76.5 \\
 Attentional Fusion & 27.6 & 17.7 & 12.2 & 8.7 & 11.7 & 29.3 & 78.7  \\
 Modified Attn. Fusion & 27.7 & 17.6 & 12.0 & 8.5 & 11.8 & 29.0 & 76.5 \\
 \hspace{3mm}+MMI objective & 28.3 & 18.1 & 12.4 & 8.9 & 12.1 & 29.6 & 80.5 \\
 Hierarchical Attention & 29.1 & 18.6 & 12.6 & 9.0 & 12.7 & 30.1 & 82.4 \\
 \hspace{3mm}+pre-trained embedding & \textbf{30.7} & \textbf{20.4} & 14.4 & 10.6 & 13.6 & 32.0 & 99.5 \\
 Multi-Source Seq2Seq & - & - & - & 10.58 & \textbf{14.13} & 36.54 & 105.39 \\
 Ours & 30.52$\pm$0.34 & 20.00$\pm$0.20 & \textbf{14.46}$\pm$0.14 & \textbf{10.93}$\pm$0.11 & 13.87$\pm$0.10 & \textbf{36.62}$\pm$0.23 & \textbf{113.28}$\pm$1.37\\
 \bottomrule
\end{tabular}
\end{adjustbox}
\caption{Comparison with existing approaches: Na\"ive Fusion~\cite{alamri@DSTC7, zhuang2019investigation}, Attentional Fusion~\cite{hori2018end, zhuang2019investigation}, Multi-Source Sequence-to-Sequence model~\citep{pasunuru2019dstc7}, Modified Attentional Fusion with Maximum Mutual Information objective~\cite{zhuang2019investigation} and Hierarchical Attention with pre-trained embedding~\citep{le2019end}, on the AVSD public test set. For each approach, we report its corpus-wide scores on BLEU-1 through BLEU-4, METEOR, ROUGE-L and CIDEr.
We report the mean and standard deviation scores of 5 runs using random initialization and early stopping on the public (prototype) validation set.
}
\label{tab:combined_res}
\end{table*}

In Table~\ref{tab:dstc7_stat}, we present the statistics of the AVSD dataset. Given the fact that the lengths of the I3D-RGB frame feature sequences are more than 20 times longer than the questions, using a recurrent neural network to encode the visual feature sequences will be very time consuming, as the visual frames are processed sequentially. Our proposed question-guided video representation module summarizes the video sequence efficiently - aggregating the visual features by question-guided attention and weighted summation and performing gating with a question-guided gate vector, both of which can be done in parallel across all frames.

\subsection{Experimental Setup}
We implement our models using the Tensor2Tensor framework~\cite{vaswani2018tensor2tensor}. The question and dialogue context tokens are both embedded with the same randomly-initialized word embedding matrix, which is also shared with the answer decoder's output embedding. The dimension of the word embedding is 256, the same dimension to which the I3D-RGB features are transformed. All of our LSTM encoders and decoder have 1 hidden layer. Bahdanau attention mechanism ~\cite{bahdanau2014neural} is used in the answer decoder. During training, we apply dropout rate $0.2$ in the encoder and decoder cells. We use the ADAM optimizer~\cite{kinga2015method} with $\alpha=2\times 10^{-4}, \beta_1=0.85, \beta_2=0.997, \epsilon=10^{-6}$, and clip the gradient with L2 norm threshold $2.0$~\cite{pascanu2013difficulty}. The models are trained up to 100K steps with early stopping on the validation BLEU-4 score using batch size 1024 on a single GPU. During inference, we use beam search decoding with beam width 3. We experimented with word embedding dimension \{256, 512\}, dropout rate \{0, 0.2\}, Luong and Bahdanau attention mechanisms, \{1, 2\} hidden layer(s) for both encoders and the decoder. We found the aforementioned setting worked best for most models.

\section{Results}
\label{sec:results}
\subsection{Comparison with Existing Methods}
We evaluate our proposed approach using the same natural language generation evaluation toolkit NLGEval~\cite{sharma2017nlgeval} as the previous approaches.
The corpus-wide scores of the following unsupervised automated metrics are reported, including BLEU-1 through BLEU-4~\cite{Papineni02Bleu}, METEOR~\cite{banerjee2005meteor}, ROUGE-L~\cite{Lin04Rouge} and CIDEr~\cite{vedantam2015cider}.
The results of our models in comparison with the previous approaches are shown in Table~\ref{tab:combined_res}. We report the mean and standard deviation scores of 5 runs using random initialization and early stopping on the public (prototype) validation set.
We apply our model in two scenarios: single-turn and multi-turn VideoQA. The only difference is that in single-turn VideoQA, the dialogue context encoder is excluded from the model.

First we observe that our proposed multi-turn VideoQA model significantly outperforms the single-turn VideoQA model. This suggests that the additional dialogue context input can provide supplementary information from the question and visual features, and thus is helpful for generating the correct answer.
Secondly, comparing the single-turn VideoQA models, our approach outperforms the existing approaches across all automatic evaluation metrics. This suggests the effectiveness of our proposed question-guided video representations for VideoQA.
When comparing with previous multi-turn VideoQA models, our approach that uses the dialogue context (questions and answers in previous turns) yields state-of-the-art performance on the BLEU-3, BLEU-4, ROUGE-L and CIDEr metrics and competitive results on BLEU-1, BLEU-2 and METEOR. It is worth mentioning that our model does not use pre-trained word embedding or audio features as in the previous hierarchical attention approach~\cite{le2019end}.

\begin{table}[t]
\centering
\begin{adjustbox}{max width=\linewidth}
 \begin{tabular}{l c c c c}
 \toprule
 Model & BLEU-4 & METEOR & ROUGE-L & CIDEr  \\
 \midrule
 Ours & 10.94 & 13.73 & 36.30 & 111.12 \\
 -TokSumm & 10.46 & 13.49 & 35.81 & 110.08 \\
 -Gating & 10.59 & 13.64 & 36.11 & 108.51 \\
 -TokSumm-Gating & 10.06 & 13.20 & 35.35 & 104.01 \\ 
 \bottomrule
\end{tabular}
\end{adjustbox}
\caption{Ablation study on the AVSD validation set. We observe that the performance degrades when either of both of the question-guided per-token visual feature summarization (TokSumm) and feature gating (Gating) techniques are removed.}
\label{tab:ablation_study}
\end{table}

\subsection{Ablation Study and Weights Visualization}

We perform ablation experiments on the validation set in the multi-turn VideoQA scenario to analyze the effectiveness of the two techniques in the question-guided video representation module. The results are shown in Table~\ref{tab:ablation_study}.

\subsubsection{Question-Guided Per-Token Visual Feature Summarization (TokSumm)}
Instead of using token-level question representations $\mathbf{x}^\text{tok} = \{x^\text{tok}_1, \ldots, x^\text{tok}_K\}$ to generate per-token video summary $\mathbf{v} = \{v_1, \ldots, v_K\}$, we experiment with using the sentence-level representation of the question $x^\text{sen}$ as the query vector to attend over the I3D-RGB visual features to create a visual summary $v$, and use $v$ to augment each of the question tokens in the video-augmented question encoder.
\begin{align}
    & s_l = \textrm{trilinear}(x^\text{sen}, r_l) \\
    & \forall l \in \{1, \ldots, L\} \notag \\
    \mathbf{w}^\text{att} &= \textrm{softmax}(\mathbf{s}) \\
    v &= \sum_{l = 1}^L w^\text{att}_l r_l
\end{align}
We observe the performance degrades when the sentence-level video summary is used instead of the token-level video summary.

\begin{figure*}[t]
    \center
	\includegraphics[width=\textwidth]{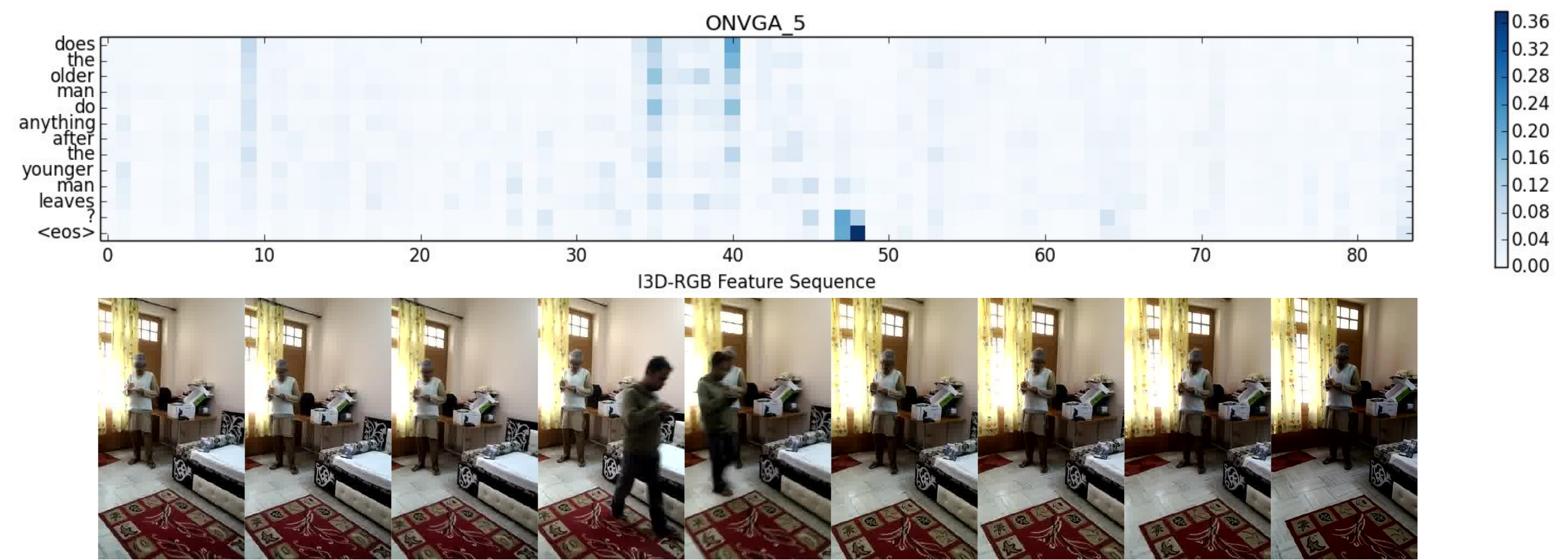}
	\caption{Question-guided per-token visual feature summary weights on a question. Each row represents the attention weights $\mathbf{w}^\text{att}_k$ of the corresponding encoded question token $x_k^\text{tok}$ over the I3D-RGB visual features. We can observe that the attention weights are shifted to focus on the relevant segment of the visual frame features for the question tokens ``after the younger man leaves \textless eos\textgreater ?"}
	\label{fig:affinity}
\end{figure*}

Figure~\ref{fig:affinity} shows an example of the attention weights in the question-guided per-token visual feature summarization. We can see that for different question tokens, the attention weights are shifted to focus on the different segment in the sequence of the video frame features.

\begin{figure*}[t]
    \center
	\includegraphics[width=\textwidth]{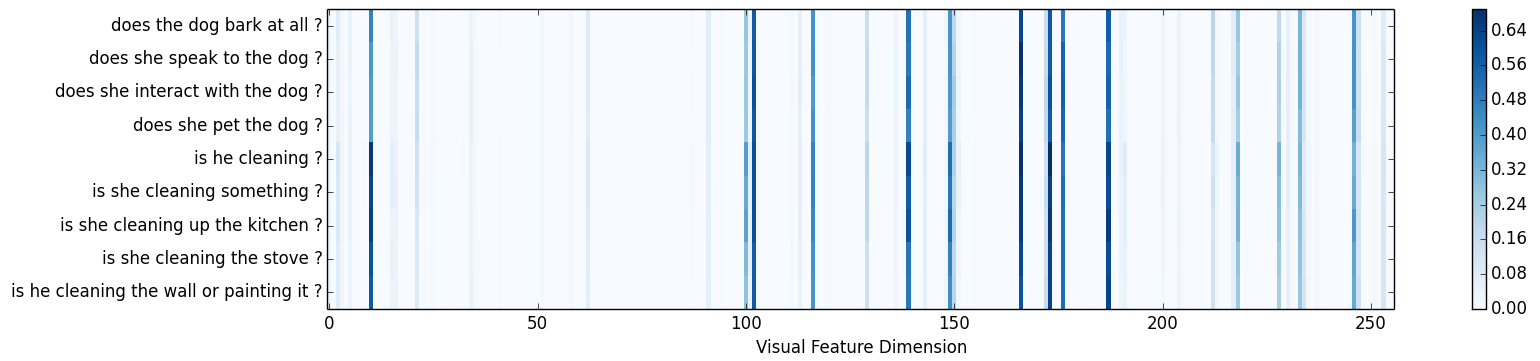}
	\caption{Question-guided gate weights $g$ for some example questions. Across the questions about similar subjects, we observe a similar trend of weight distribution over visual feature dimensions. Conversely, questions about different topics show different gate weights patterns.}
	\label{fig:gating}
\end{figure*}

\subsubsection{Question-Guided Visual Feature Gating (Gating)}
We also experiment with using the non-gated token-level video summary $\mathbf{v} = \{v_1, \ldots, v_K\}$ to augment the question information in the video-augmented question encoder.
We observe the model's performance declines when the question-guided gating is not applied on the video summary feature.
Removing both the per-token visual feature summarization and the gating mechanism results in further degradation in the model performance.

Figure~\ref{fig:gating} illustrates the question-guided gate weights $g$ of several example questions. We observe that the gate vectors corresponding to the questions regarding similar subjects assign weights on similar dimensions of the visual feature. Although many of the visual feature dimensions have low weights across different questions, the feature dimensions of higher gate weights still exhibit certain topic-specific patterns.
\section{Conclusion and Future Work}
In this paper, we present an end-to-end trainable model for single-turn and multi-turn VideoQA. Our proposed framework takes the question, I3D-RGB video frame features and dialogue context as input. Using the question information as guidance, the video features are summarized as compact representations to augment the question information, which are jointly used with the dialogue context to generate a natural language answer to the question.
Specifically, our proposed question-guided video representation module is able to summarize the video features efficiently for each question token using an attention mechanism and perform feature selection through a gating mechanism.
In empirical evaluation, our proposed models for single-turn and multi-turn VideoQA outperform existing approaches on several automatic natural language generation evaluation metrics.
Detailed analyses are performed, and it is shown that our model effectively attends to relevant frames in the video feature sequence for summarization, and the gating mechanism shows topic-specific patterns in the feature dimension selection within a frame.
In future work, we plan to extend the models to incorporate audio features and experiment with more advanced techniques to incorporate the dialogue context with the question and video information, such as hierarchical attention and co-attention mechanisms. We also plan to employ our model on TVQA, a larger scale VideoQA dataset.

\bibliography{guanlinc}
\bibliographystyle{acl_natbib}

\end{document}